\documentclass[10pt,journal,compsoc]{IEEEtran}
\usepackage[left=1.5cm,right=1.5cm,top=2cm,bottom=2cm]{geometry}
\usepackage{ragged2e}
\justifying

\ifCLASSOPTIONcompsoc
   \usepackage[nocompress]{cite}
\else
    \usepackage{cite}
\fi

\ifCLASSINFOpdf
 
\else
 
\fi
\usepackage{multirow}
\usepackage{graphicx}
\usepackage{xcolor}
\usepackage{tabularx}

\usepackage{authblk}

\begin{document}

\title{Automatic summarisation of Instagram social network posts
Combining semantic and statistical approaches}

\author{\IEEEauthorblockN{Kazem Taghandiki\IEEEauthorrefmark{1},
                        Mohammad Hassan Ahmadi\IEEEauthorrefmark{2},
                        Elnaz Rezaei Ehsan\IEEEauthorrefmark{3}}
	\IEEEauthorblockA{
	 \\
		\IEEEauthorrefmark{1}Department of Computer Engineering, Technical and Vocational University (TVU), Tehran, Iran
 \\ktaghandiki@tvu.ac.ir
 \\
		\IEEEauthorrefmark{2} Department of Electrical Engineering, Technical and Vocational University (TVU), Tehran, Iran\\  ahmadi.m4942@gmail.com  \\
  		\IEEEauthorrefmark{3} Master's degree, industrial engineering, system management and productivity, Iran University of Science and Technology\\  elnazrezaeie110@gmail.com  \\
                            }
							}
\IEEEtitleabstractindextext{%
\begin{abstract}
\textcolor{black}{The proliferation of data and text documents such as articles, web pages, books, social network posts, etc. on the Internet has created a fundamental challenge in various fields of text processing under the title of "automatic text summarisation". Manual processing and summarisation of large volumes of textual data is a very difficult, expensive, time-consuming and impossible process for human users. Text summarisation systems are divided into extractive and abstract categories. In the extractive summarisation method, the final summary of a text document is extracted from the important sentences of the same document without any modification. In this method, it is possible to repeat a series of sentences and to interfere with pronouns. However, in the abstract summarisation method, the final summary of a textual document is extracted from the meaning and significance of the sentences and words of the same document or other documents. Many of the works carried out have used extraction methods or abstracts to summarise the collection of web documents, each of which has advantages and disadvantages in the results obtained in terms of similarity or size. In this work, a crawler has been developed to extract popular text posts from the Instagram social network with appropriate preprocessing, and a set of extraction and abstraction algorithms have been combined to show how each of the abstraction algorithms can be used. Observations made on 820 popular text posts on the social network Instagram show the accuracy (80$\%$) of the proposed system.. }
\end{abstract}

}

\maketitle

\IEEEdisplaynontitleabstractindextext

\IEEEpeerreviewmaketitle

\section{Introduction}
The ever-increasing amount of data and text documents such as articles, web pages, books, social network posts, etc. on the Internet has created a fundamental challenge in various fields of text processing under the title of "automatic text summarisation" \cite{b1}. Text summarisation systems are widely used in text processing and analysis applications such as information retrieval, information mining, and question and answer systems \cite{b2}. Manual processing and summarisation of large amounts of textual data is a very difficult, costly, time-consuming and impossible process for human users \cite{a1,b3}. The main purpose and application of text summarisation systems is to produce short and abstract text from important sentences of a set of input documents \cite{b4}. Text summarisation systems allow users to have faster access to information in input documents without having to read them all \cite{a2,b5}. The input to a text summarisation system can be single document or a multi-document. In single-document mode, the system creates a brief summary of important sentences in only one document, but in multi-document mode, the system creates a summary from thousands of documents \cite{a3}. Text summarisation systems are divided into extractive and abstract categories. In extractive summarisation, the final summary of a document is extracted from the important sentences of the same document without any changes. In this method, it is possible to repeat a series of sentences and interfere with pronouns. However, in the abstract summarisation method, the final summary of a document is extracted from the meaning and significance of the sentences and words of the same document or other documents \cite{b1}.\\
Many of the performed works have used extractive or abstract methods for summarising the collection of web documents, each of which has advantages and disadvantages in the degree of similarity \cite{a4,a6,a5}.
Many of the performed works have used extractive or abstract methods to summarise the collection of web documents, each of which has advantages and disadvantages in the degree of similarity and the size of the summary produced. In the proposed approach, the researcher has used a combined method (abstract and extractive) to summarise the textual data of many viewers in the social network Instagram. The observations show the acceptable accuracy of the proposed approach.
\subsection{Automatic text summarisation systems}
ATS or automatic text summarization systems is one of the fundamental challenges in the field of natural language processing and artificial intelligence. The first research on this topic was done by Mr. Len in 1958 to extract text summaries from articles \cite{b6}. Radeff \cite{b7} and his colleagues, by continuing the research and work done, solved the challenge of how to identify the important parts of a document. Han and Mani \cite{b8} solved the challenge of using large documents such as books in automatic text summarization systems. The main idea of ATS systems, or automatic text summarization, is to generate a short summary of input documents in a smaller volume \cite{b1}, which helps users to extract the main idea and topic of large text documents without having to read the entire text. Radeff \cite{b7} defines summarization as extracting text from one or more large input documents whose size is at most half the size of the input documents. Gambhir \cite{b9} defines summarization as a process in which the size of the final text produced is much smaller than the size of the input text documents. The general architecture of an automatic text summarization system is shown in Figure 1.\\
   \begin{figure}
    \centering
    \includegraphics[width=9cm,height=2cm]{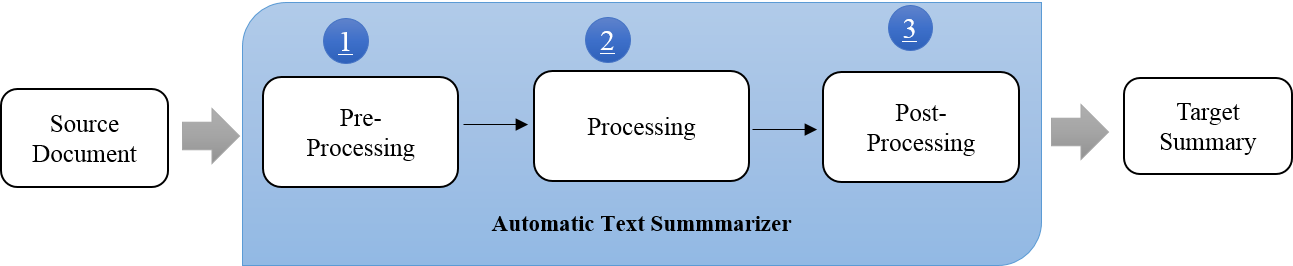}
    \caption{The general architecture of an automatic text summarisation system \cite{b1}}
    \label{fig:life}
\end{figure}
In the architecture of Figure 1, a set of documents called the corpus, is first provided to the ATS system in the form of an input. The ATS system then performs the task of automatically summarizing the incoming documents in three phases.\\
\textbf{1. Pre-processing phase:} The purpose of this phase is to convert the unstructured text of the input documents into a quality structure. Typically, various linguistic methods and tools are used for this task, such as removing stop words, recognising the existence of a word and finding roots \cite{a7,b1}.\\
\textbf{2. Processing phase:} In this phase, different document summarisation approaches, such as abstract, extractive and combined approaches, are used to classify the sentences of all the documents pre-processed in the previous phase \cite{a10,b1}.\\
\textbf{3. Post-processing  phase:} In this phase, the existing relationship between the sentences and their importance is discovered, finally by combining a set of sentences, the final summary is formed \cite{a8,b1}.\\

In the continuation of this article, in section 2, the author describes some of the work done for different text summarisation methods. In section 3, the author will implement the proposed approach of automatic text summarisation system. In section 4, the author will review the results and observations obtained from the proposed approach, and in section 5, the final conclusion will be discussed.
\section{Related Work}
The field of automatic summarization has been the subject of extensive research for several decades, but with the explosive growth of social media platforms, summarizing social network posts has become a particularly challenging task \cite{a9,a11}. Instagram, as one of the most popular social networks, generates vast amounts of textual content daily, making it an ideal candidate for automatic summarization. This paper presents a novel approach to summarizing Instagram social network posts by combining semantic and statistical methods. The proposed approach leverages natural language processing techniques to extract semantic information from the posts while also utilizing statistical methods to identify the most important sentences. This section will review the existing literature on automatic summarization techniques and their application to social network posts and will then describe the proposed approach in detail.

Aggarwal et al \cite{b10} used extractive methods to summarise the textual feedback of videos available on the web. In the first phase, they collected and pre-processed the users' textual feedback about the videos, in the pre-processing process they used the methods of stop word removal and lexical rooting to improve the basic structure of the texts. In the second phase, they carried out the feature extraction process on the pre-processed texts from the previous phase; in fact, in this phase, each text document was represented as a feature vector of unigrams, bigrams and trigrams of the words present. In the third phase, using the naive Bayesian algorithm, each of the text documents (video text feedback) was evaluated with positive and negative labels indicating the positive and negative feedback of the users towards a video. Finally, in the fourth stage, the sentences in the feedback were grouped and weighted using the k-means clustering method to finally extract the most important sentences as a summary of a text document.\\
Vangara et al \cite{b11} used extractive methods to build a document summarisation system. Their system was divided into 3 parts, in the first part, the corpus of TIPSTER documents with 180 documents was given as input to the system. In the second part, the documents and texts were first preprocessed (removal of stop words and root search) using the Weka tool, then the sentences were weighted with They using a Java program according to the TFIDF statistical analysis and the semantic similarity between the words. Finally, in the third part, the system returns the most important sentences as a summary of the input document.\\
Neto et al \cite{b12} used abstract summarisation or semantic methods to identify important sentences in a document. First, they performed pre-processing operations (such as removing stop words, rooting) on the sentences of a document, and then identified the names and concepts in the document using the post of tagger tools. The next step was to examine each of the identified names in the wordnet dictionary. Each concept or name in the wordnet dictionary has a coherence and correlation with other concepts and names. Finally, important sentences from a document were selected according to the factors of coherence and correlation of concepts.\\
Khan et al \cite{b13} proposed a 3-stage system for summarising texts. In the first stage, the system used a crawler to collect the textual content of the news pages of the website and stored it in a scrap. In the second phase, the system first performed pre-processing (removal of stop words and stemming) on the documents, and then clustered the documents using statistical parameters such as TFIDF. Finally, in the third phase, a similarity matrix was extracted and weighted for each sentence of the documents, showing the similarity of each sentence to each other, and the important sentences of each document were identified and extracted according to the sentence rank \cite{a12}.
\section{The proposed approach}
In this section, the researcher has implemented the proposed approach according to the process of Figure 2. Each step of the proposed approach in Figure 2 is described below.
   \begin{figure}
    \centering
    \includegraphics[width=9cm,height=4cm]{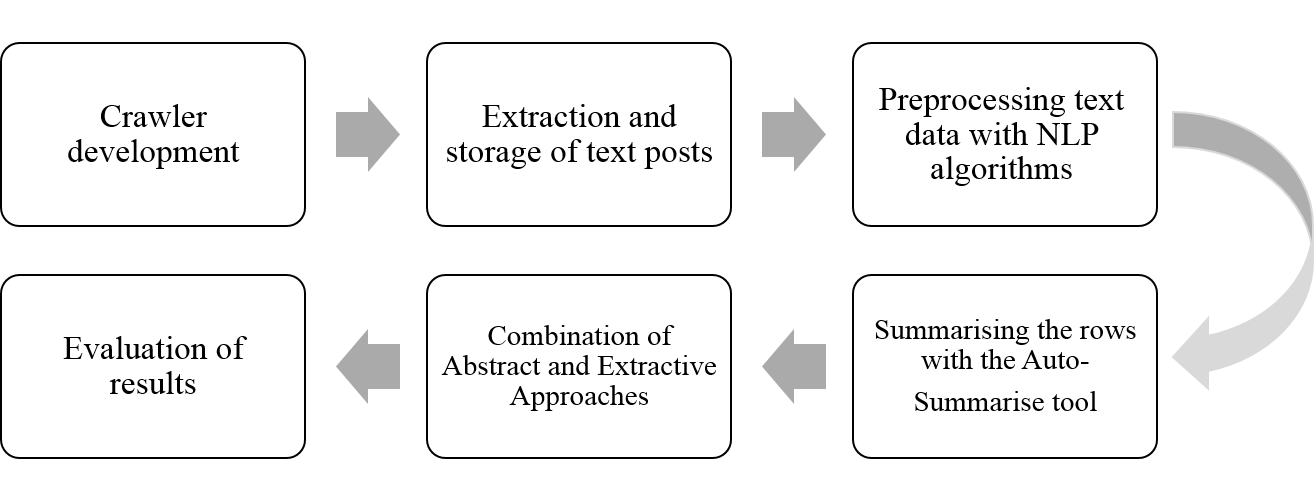}
    \caption{Process of the proposed approach}
    \label{fig:life}
\end{figure}

\subsection{Crawler development}
To implement and evaluate the proposed approach, a dataset of text posts on the social network Instagram through popular hashtags is needed. Therefore, a crawler was first developed with the help of BeautifulSoup and Scrapy library in the Google Colab environment, the developed crawler extracted the collection of posts on Instagram pages that had a special and popular hashtag and saved it in the form of a CSV file. Table 1 shows some of the popular and trending hashtags of the Instagram social network.

\begin{table}[]
\tiny
\centering
\caption{Popular and trending hashtags on the Instagram social network}
\label{tab:my-table}
\begin{tabular}{|c|c|c|c|}
\hline
Number of times used & Popular hashtag & Number of times used & Popular hashtag \\ \hline
(1.835B)             & \#love          & (578.8M)             & \#happy         \\ \hline
(1.150B)             & \#instagood     & (570.8M)             & \#picoftheday   \\ \hline
(812.7M)             & \#fashion       & (569.1M)             & \#cute          \\ \hline
(797.3M)             & \#photooftheday & (560.9M)             & \#follow        \\ \hline
(661.0M)             & \#beautiful     & (536.4M)             & \#tbt           \\ \hline
(649.9M)             & \#art           & (528.5M)             & \#followme      \\ \hline
(583.1M)             & \#photography   & (525.7M)             & \#nature        \\ \hline
\end{tabular}
\end{table}

\subsection{Extraction and storage of text posts}
Once the popular hashtags have been selected, they can be provided to the crawler to extract the posts that contain that hashtag. Finally, with the crawler developed in this section, approximately 886 Instagram text posts with popular hashtags were examined and extracted, each of which was stored in a row of the csv file named dataset.

\subsection{Preprocessing text data with NLP algorithms}
After saving the text posts extracted from the social network Instagram in separate lines of the dataset file, it was found that each line contained a large number of unprintable and inappropriate characters (! $\$$( ) * $\%$ @), which were removed by removing them. Finally, we can achieve better accuracy in the next steps. In addition, natural language preprocessing such as word separation, stop word removal, normalisation and duplicate line removal were performed on the data set file. Table 2 shows the result of the volume and character analysis after applying preprocessing using natural language processing libraries such as nltk.

\begin{table}[]
\tiny
\centering
\caption{The effect of pre-processing on the dataset}
\label{tab:my-table}
\begin{tabular}{|cc|cc|}
\hline
\multicolumn{2}{|c|}{Dataset file specifications after preprocessing} & \multicolumn{2}{c|}{Dataset file specifications before preprocessing} \\ \hline
\multicolumn{1}{|c|}{The file size is}               & 4 MB           & \multicolumn{1}{c|}{The file size is}                & 5 MB           \\ \hline
\multicolumn{1}{|c|}{Number of characters}           & 200125         & \multicolumn{1}{c|}{Number of characters}            & 258855         \\ \hline
\multicolumn{1}{|c|}{The number of lines is}         & 820            & \multicolumn{1}{c|}{The number of lines is}          & 886            \\ \hline
\end{tabular}
\end{table}

\subsection{Summarising the rows of the dataset with the AutoSummarise tool}
In order for the researcher to be able to obtain the values of the evaluation parameters of the proposed model, each line of the data set file must be summarised with an accurate and popular tool, so that at the end, by comparing the outputs, he can check and analyse the results. Therefore, the lines of the pre-processed data set file were summarised using the well-known tool AutoSummarize. AutoSummarize is an intelligent text summarisation tool that comes with Microsoft Word software, so researchers can use it to perform the process of summarising text documents in Word and to validate and evaluate their proposed approaches. Figure 3 shows an example of a summary produced by this intelligent tool. This process was performed on all the remaining 820 lines of the dataset file over 2 days by 5 expert users. Finally, in addition to the pre-processed dataset file with 820 lines (text posts from the social network Instagram), an output file named automatic summary with the same number of lines (each summary line corresponds to one line in the dataset file) was saved with the extension csv.
\begin{figure*}[h!]
    \centering
    \includegraphics[width=14cm,height=5cm]{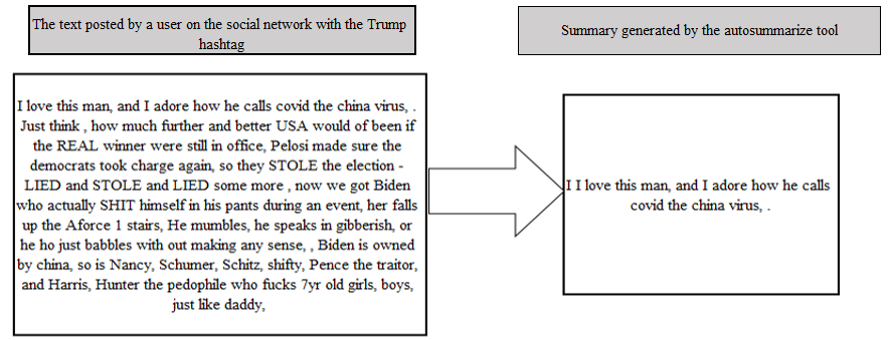}
    \caption{An example of the summary generated by the auto summarize smart tool}
    \label{fig:life}
\end{figure*}

Table 3 shows the volume and character result between the two pre-processed dataset files and the autosummarize file.

\begin{table}[]
\tiny
\centering
\caption{Comparison of dataset and autosummarize file}
\label{tab:my-table}
\begin{tabular}{|cc|cc|}
\hline
\multicolumn{2}{|c|}{autosummarize file specifications} & \multicolumn{2}{c|}{Dataset file specifications}     \\ \hline
\multicolumn{1}{|c|}{The file size is}         & 2 MB   & \multicolumn{1}{c|}{The file size is}       & 4 MB   \\ \hline
\multicolumn{1}{|c|}{Number of characters}     & 88256  & \multicolumn{1}{c|}{Number of characters}   & 200125 \\ \hline
\multicolumn{1}{|c|}{The number of lines is}   & 820    & \multicolumn{1}{c|}{The number of lines is} & 820    \\ \hline
\end{tabular}
\end{table}

\subsection{ Combination of Abstract and Extractive Approaches}
At this stage, by combining abstract and extractive summarisation algorithms, the researcher intends to create a text summarisation system with the highest percentage of similarity and accuracy compared to the automatic summarisation tool. To develop the extracting part of the automatic text summarisation system, two algorithms textrank and lexrank have been used to generate a summary of each line of the data set file, after each of the two algorithms textrank and lexrank has made its proposed summary for each line of the data set provided. Between the two generated summaries, the summary that is more similar to the input text and has a smaller volume than the summary of the other algorithm is selected \cite{b15}.\\
\textbf{1. textrank and lexrank algorithms}\\
The textrank algorithm selects important sentences from a text based on the concept, meaning and frequency of words. lexrank algorithm is the second algorithm chosen for the development of the extraction part of the proposed automatic summarisation system. This algorithm selects, from the set of sentences of the input or original text, the sentence that has the highest percentage of similarity with other sentences of the input text.
In the same way, to develop the abstract part of the proposed system, the T5 and BART algorithms were used to produce a summary of each line of the pre-processed dataset file. For each line of the dataset file, one of the two generated summaries is selected that has more similarity to the input text and less volume than the summary of the other algorithm  \cite{a15,b15}.\\
\textbf{T5 and BART algorithm} \\
The t5 algorithm is an encoder-decoder model that receives its input as a sequence of codes or in other words encoding. So after extracting a word using the t5 algorithm, the user must convert or encode his input into a sequence of codes. Then by decoding the sequence of codes generated in the previous step, the user can form the final summary using the t5 algorithm. The BART algorithm, like the T5 abstract algorithm, requires two stages of encoding and decoding. In the encoding step, each line of the pre-processed data set file is converted into a sequence of codes, and in the decoding step, the trained bart-large-cnn model is used to summarize each line of the pre-processed data file. to be \cite{b16,a13}.\\
As explained above, each of the extraction and abstraction approaches has two different algorithms. Among the outputs (summaries) produced by each of the algorithms, the best summary is selected that is more similar to the input text and the number of words produced in the summary. Thus, in the end, the system has two summaries generated for each input sentence or line from the data set file. One of the summaries is related to the extractive part of the system and the other is related to the abstract part \cite{a14}. Therefore, again from the two summaries produced from the extractive and abstract parts, the best summary that is more similar to the input text and has fewer words in the produced summary is selected. Figure 4 shows the final model of the developed text summarisation system.
\begin{figure*}[h!]
    \centering
    \includegraphics[width=12cm,height=6cm]{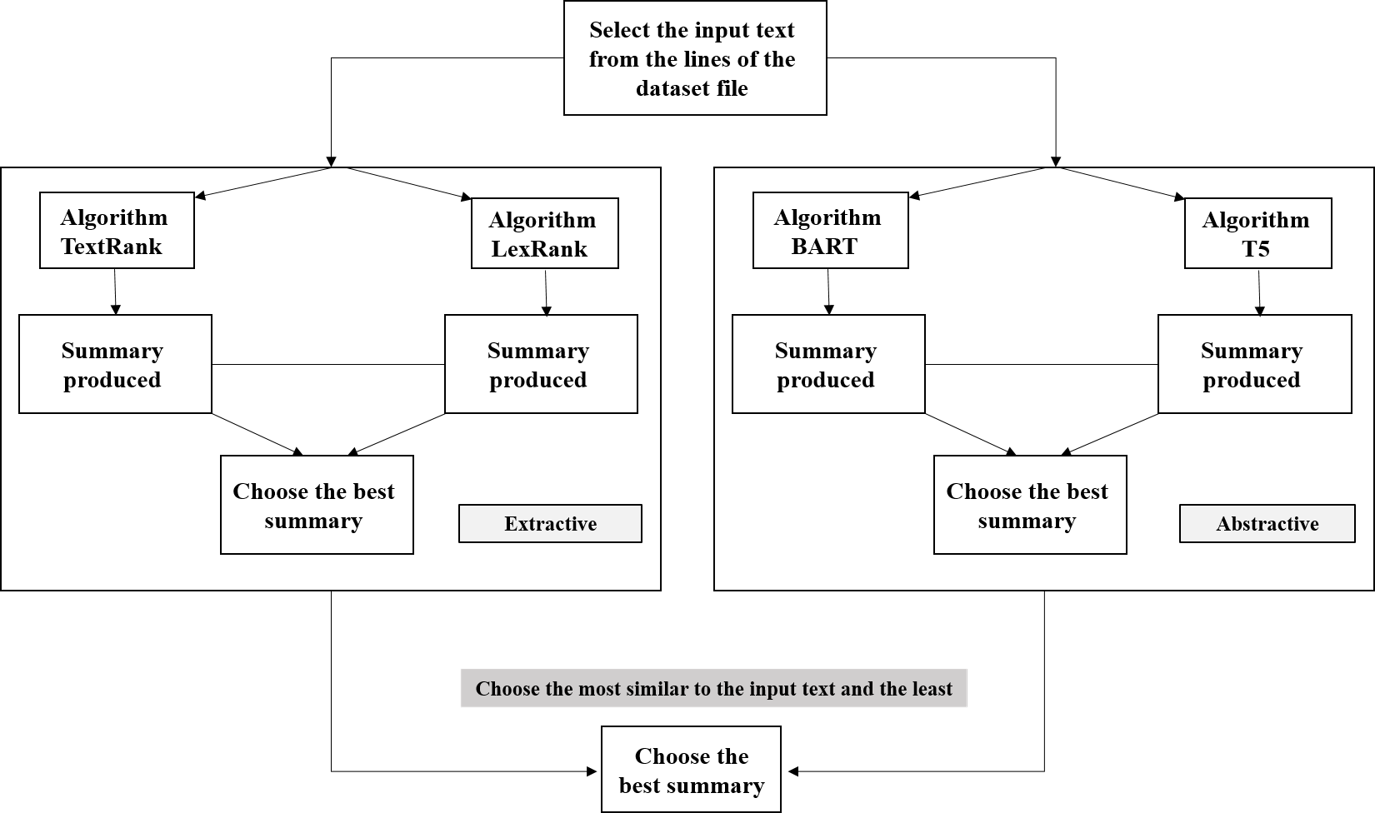}
    \caption{Combined model of deductive and abstract approaches}
    \label{fig:life}
\end{figure*}
In all stages of selecting the most similar summary to the lines of the dataset file, the combination of the semantic similarity of the WS4J library \cite{b17} and the cosine string \cite{b18} has been used. The researcher has studied the last stage of the proposed approach in section 4 of this article.
\section{observations}
The purpose of this part is to verify, evaluate and compare the results of the developed text summarisation system with the results of the automatic summarisation tool, which will be explained in detail later. According to the final model developed in section 3, Figure 5 shows the number of selections of the best final summary among the summaries generated from the two extractive and abstract sections.
\begin{figure}
    \centering
    \includegraphics[width=9cm,height=5cm]{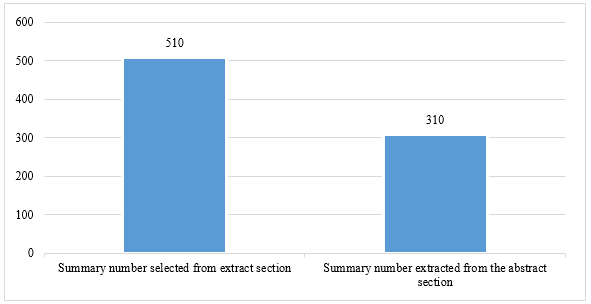}
    \caption{The number of final summaries selected from the two extractive and abstract parts of the developed system}
    \label{fig:life}
\end{figure}
As shown in Figure 5, of the 820 summaries obtained, approximately 510 summaries were selected by the extractive summarisation section and 310 summaries were selected by the abstract summarisation section. Therefore, about 510 of the results or summaries were obtained by extractive approaches and 310 by abstract approaches. This shows that the extraction part of the developed system has produced a better final summary in terms of semantic and string similarity with the dataset file.\\
Another essential analysis to be studied is the similarity of the final summaries produced by the developed text summarisation system with the summaries produced by autosummarize, as shown in Figure 6.
\begin{figure}
    \centering
    \includegraphics[width=9cm,height=5cm]{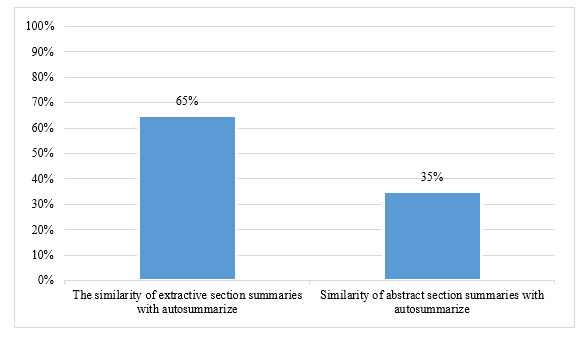}
    \caption{The similarity of the summaries produced by the two extractive and abstract parts of the system with the autosummarize tool}
    \label{fig:life}
\end{figure}
As shown in Figure 6, the 510 summaries produced by the extraction section have a similarity of about 65$\%$ with the summaries produced by the autosummarise tool, and the 310 summaries produced by the abstract section have a similarity of about 35$\%$ with the summaries produced by the autosummarise tool.Therefore, according to the results obtained, the extractive summarisation part of the system can be considered similar to the autosummarise tool.\\
Now, if the summaries produced by the autosummarize tool are checked by an expert user and the correct summary is marked with the label P and the wrong summary (or in other words, it could be better) is marked with the label N, it is easy to obtain the confusion matrix and finally the information retrieval evaluation criteria such as accuracy, recall, precision and f-measure. Therefore, each summary produced by the autosummarize tool was reviewed by 3 users and in 3 days, and P and N labels were assigned to them. Figure 7 shows the confusion matrix according to correct labels (summaries produced by the autosummarize tool) and predicted labels (summaries produced by the developed system).
\begin{figure}
    \centering
    \includegraphics[width=5cm,height=2cm]{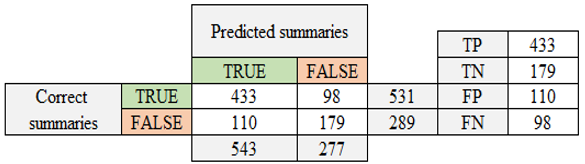}
    \caption{The extracted confusion matrix}
    \label{fig:life}
\end{figure}
\\To generate the confusion matrix, the researcher used a threshold of 80$\%$ similarity between the summary generated by the developed system and the summary obtained by the automatic summarisation tool for each line of the dataset file, so that if the summary generated for each line of the dataset file is at least 80$\%$ similar to the summary obtained by the automatic summarisation tool, it is assigned the label P, otherwise the label N. At the end of this process, the confusion matrix shown in Figure 7 was obtained.\\
\textbf{TP:} indicates the number of summaries correctly obtained by the automatic summarisation tool and the automatic summarisation system.\\
\textbf{TN:} indicates the number of summaries that could not be extracted by the automatic summarisation tool and the automatic summarisation system.\\
\textbf{FP:} Number of summaries not detected by the autosummarise tool but detected by the automatic summarisation system.\\
\textbf{FN:} Indicates the number of summaries that were not incorrectly identified by the autosummarise tool and the automatic summarisation system.\\
Now, according to the values of the variables TP, FP, TN and FN, the evaluation criteria can be easily obtained \cite{g1,g2}. Figure 8 shows the statistical results of the evaluation criteria of accuracy, recall, precision, f-measure and error rate.
\begin{figure}
    \centering
    \includegraphics[width=7cm,height=4cm]{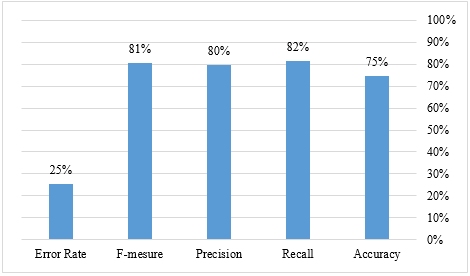}
    \caption{Statistical results of accuracy, recall, precision, f-measure and error rate evaluation criteria}
    \label{fig:life}
\end{figure}
As shown in Figure 8, the accuracy, completeness, precision, harmonic mean and error rates of the developed automatic text summarisation system have values of 75$\%$, 82$\%$, 80$\%$, 81$\%$ and 25$\%$ respectively, which shows the optimal and ideal accuracy of the developed system.
\section{Conclusion}
Text summarisation systems are divided into extractive and abstract categories. In the extractive summarisation method, the final summary of a text document is extracted from the important sentences of the same document without any kind of modification, in this method it is possible to repeat a series of sentences in a repetitive form and interfere with pronouns. However, in the abstract summarisation method, the final summary of a textual document is extracted from the meaning and significance of the sentences and words of the same document or other documents \cite{b1}. Many of the performed works have used extraction methods or abstracts to summarise the collection of web documents, each of which has advantages and disadvantages in the results obtained in terms of similarity or size. In this work, we have developed a crawler that extracts popular text posts from the social network Instagram, with appropriate preprocessing and we have combined a set of extractive and abstractive algorithms. Observations on 820 popular text posts on the social network Instagram show the accuracy (80$\%$) and precision (75$\%$) of the proposed system.

\section{Future works}
The proposed system for the automatic summarisation of popular text posts on Instagram social network has demonstrated promising results. However, there is scope for future work in this area to further improve the accuracy and efficiency of the summarisation process. One potential direction for future work could be to explore the use of deep learning techniques, such as neural networks, to enhance the performance of the summarization system. Another avenue for improvement could be to investigate the use of domain-specific knowledge, such as topic modeling or sentiment analysis, to improve the quality and relevance of the extracted summaries \cite{a16}. Finally, the system could also be extended to support summarization of multimedia content, such as images and videos, to provide a more comprehensive summarisation of social media content.

\bibliographystyle{IEEEtran}
\bibliography{References}

\end{document}